\documentclass[letterpaper, 10 pt, conference]{ieeeconf}  

\IEEEoverridecommandlockouts                              

\usepackage{graphicx} 
\usepackage{amsmath} 
\usepackage{amssymb}
\usepackage{multirow}
\usepackage[inkscapelatex=false]{svg}
\usepackage{adjustbox}
\usepackage{siunitx}
\usepackage[skip=2pt]{caption}

\usepackage{cite}
\usepackage{algorithm}
\usepackage{algpseudocode}
\usepackage{cuted}
\usepackage{comment}
\usepackage{bm}

\title{\LARGE \bf
Dynamic Modeling and Efficient Data-Driven Optimal Control for Micro Autonomous Surface Vehicles}

\author{Zhiheng Chen$^{1,2}$ and Wei Wang$^1$
\thanks{$^{1}$Marine Robotics Lab, Department of Mechanical Engineering, College of Engineering, University of Wisconsin-Madison.}
\thanks{$^{2}$Sibley School of Mechanical and Aerospace Engineering, Cornell University.}
\thanks{
$^{\ast}$Corresponding author: {\tt\small wwang745@wisc.edu}.}
}

\begin{document}

\maketitle
\thispagestyle{empty}
\pagestyle{empty}

\begin{abstract}
Micro Autonomous Surface Vehicles (MicroASVs) offer significant potential for operations in confined or shallow waters and swarm robotics applications. However, achieving precise and robust control at such small scales remains highly challenging, mainly due to the complexity of modeling nonlinear hydrodynamic forces and the increased sensitivity to self-motion effects and environmental disturbances, including waves and boundary effects in confined spaces. This paper presents a physics-driven dynamics model for an over-actuated MicroASV and introduces a data-driven optimal control framework that leverages a weak formulation-based online model learning method. Our approach continuously refines the physics-driven model in real time, enabling adaptive control that adjusts to changing system parameters. Simulation results demonstrate that the proposed method substantially enhances trajectory tracking accuracy and robustness, even under unknown payloads and external disturbances. These findings highlight the potential of data-driven online learning-based optimal control to improve MicroASV performance, paving the way for more reliable and precise autonomous surface vehicle operations.
\end{abstract}
\section{Introduction}
Micro Autonomous Surface Vehicles (MicroASVs), with their centimeter-scale dimensions, offer significant advantages for operations in confined or shallow waters, where larger ASVs struggle to maneuver~\cite{rudnick2004,schill2017,spino2024}. Moreover, the low cost of MicroASVs makes them well-suited for swarm robotics 
\cite{du2023,wang2020,paulos2015,o2014,furno2017},
enabling cooperative multi-robot tasks such as environmental monitoring, exploration, and surveillance. However, achieving precise and robust control at this scale is challenging due to the complexity of hydrodynamic interactions and the increased susceptibility to self-motion effects and environmental disturbances, including currents, waves, boundary effects in confined spaces, and the movements of nearby robots.

Given the complexity of hydrodynamic interactions, model-based controllers offer a promising solution for ASVs, particularly for MicroASVs, which pose greater control challenges due to their low mass and inertia. However, many existing MicroASVs continue to rely on heuristic or simplified control strategies rather than on rigorously developed dynamics models \cite{kawamura2019,kawamura2021}. Consequently, these simplified approaches often fail to capture the intricate hydrodynamic interactions in their operating environments, resulting in suboptimal trajectory tracking and reduced robustness, especially under external disturbances or payload variations \cite{han2016,muske2008}. Moreover, while larger vehicles can tolerate minor modeling inaccuracies, MicroASVs are highly sensitive to even slight discrepancies, potentially leading to significant trajectory deviations. Therefore, the development of a systematic model identification methodology is imperative for deriving an accurate dynamic model that ensures both precise system representation and reliable control performance.

Moreover, the inherently low mass and inertia of MicroASVs make them particularly vulnerable to both self-induced motion effects and external disturbances. Even minimal control inputs may lead to substantial deviations, while environmental forces—such as waves, currents, and interactions with nearby robotic platforms—can easily surpass the available actuation forces. These challenges emphasize the necessity for optimal control strategies that can robustly manage model uncertainties to ensure stable and efficient navigation. Currently, many ASV controllers rely on PID or other non-optimal approaches \cite{knizhnik2020,knizhnik2022}. Although optimal controllers, such as Model Predictive Control (MPC), have demonstrated excellent performance, their substantial computational demands render them less suitable for MicroASVs, which typically operate with limited computing resources \cite{wang2021,wang2024,wang2018}. Given these limitations, efficient optimal controllers based on linearization techniques—such as Linear Quadratic Regulators (LQR) \cite{feng2021}—offer a promising alternative. However, although LQR controllers have demonstrated effective performance in localized settings, they often struggle to account for system uncertainties and external disturbances. Consequently, there is a pressing need for more adaptive or learning-based optimal control strategies.

Building on the preceding discussion, this paper presents a physics-driven dynamics model for a MicroASV and introduces a data-driven optimal control framework that leverages a weak formulation-based online model learning method. In contrast to previous research in weak-form parameter estimation for autonomous systems \cite{bortz2023,messenger2021,bramburger2024}, our method explicitly incorporates control inputs to identify forcing terms within the governing equations, thereby enhancing model fidelity and control effectiveness. Furthermore, our optimal control problem is formulated as a two-point boundary value problem (TPBVP) based on variational principles, which can be efficiently solved and is therefore suitable for implementation on microASVs with limited computing power. The framework facilitates online learning of real-time dynamics using a samll set of moving horizon historical robott state and control action, enabling continuous model refinement to accommodate evolving system parameters and external disturbances.
The principal contributions of this work are: 
\begin{itemize}
 \item An efficient data-driven optimal control strategy for MicroASVs that leverages the real-time learned dynamics to enhance the control accuracy and robustness. 
\item \noindent  A physics-driven dynamics model for MicroASVs, which captures the drag force, the added mass effect, and the Coriolis and centrifugal effects. 
 \item \noindent A weak formulation-based algorithm for online dynamics learning. 
\end{itemize}

Extensive simulation experiments validate the proposed approach, demonstrating significant improvements in trajectory tracking accuracy and robustness, even in the presence of unknown payloads and external disturbances.

\section{Physics-Driven Dynamics Model}
\label{sec: physics-driven model}
An accurate dynamics model is critical for developing optimal controllers, particularly for micro robots that are challenging to control due to their inherently low mass and inertia. We develop a physics-driven dynamics model for MicroASVs based on Lagrange's equations in this section.
\subsection{Overview of the MicroASV Prototype}
\begin{figure}[!htb]
\centering
\includegraphics[width=0.5\textwidth]{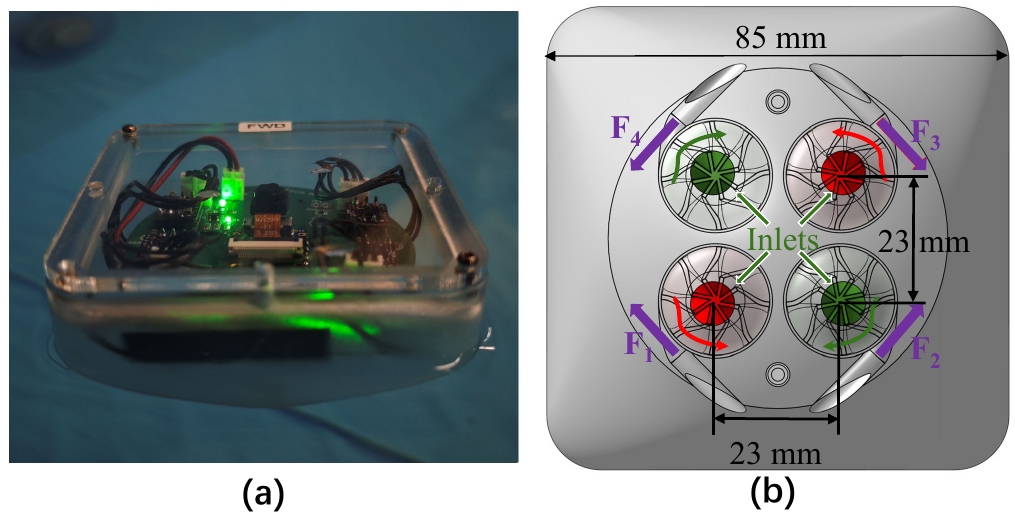}
\caption{MicroASV prototype. (a) MicroASV navigating on water; (b) Bottom view displaying the thruster configuration.}
\label{fig: prototype}
\end{figure}

The MicroASV \cite{macauley2025} we used in this work is illustrated in Fig. \ref{fig: prototype}, the MicroASV features a compact square design with an 85 mm side length.
The propulsion system consists of four custom water jets arranged in a ``diamond" configuration, driven by magnetically coupled brushless motors. This design enables full 2D mobility, including forward and backward motion, lateral translation, and in-place rotation. Constructed using commercially available motors and 3D-printed components, the MicroASV features a modular, appendage-free structure for easy assembly. Additionally, an onboard camera and inertial measurement unit (IMU) facilitate real-time localization. 
\subsection{Coordinates and General Form of Motion}
\begin{figure}
    \centering
    \includegraphics[width=0.35\textwidth]{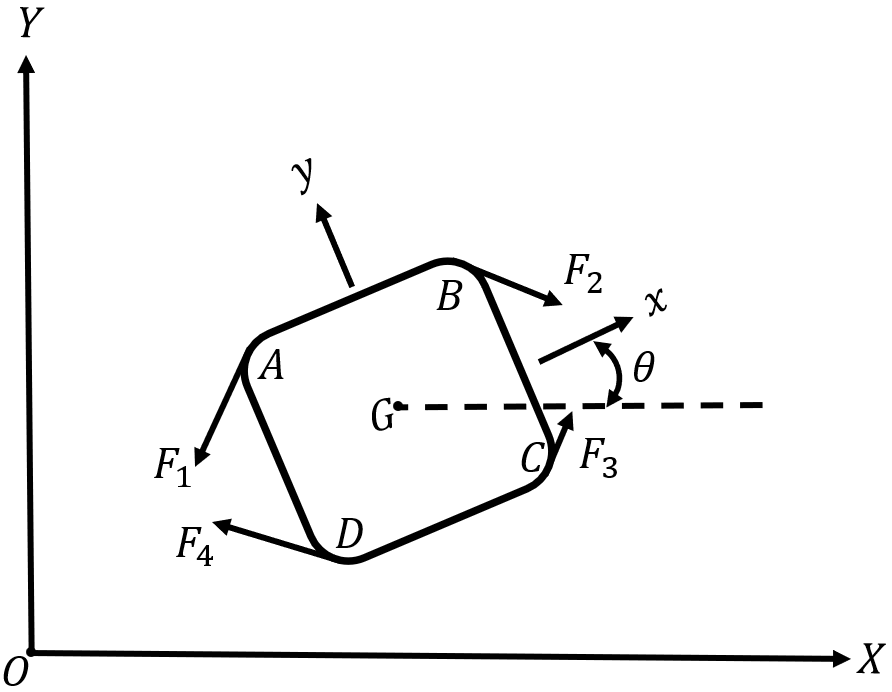}
    \caption{MicroASV coordinate systems and thruster forces. $XYZ$ denotes the inertial frame, and $xyz$ denotes the body-fixed frame. Point $O$ is the origin of the inertial frame, and point $G$ is the center of mass of MicroASV.}
    \label{fig: C-SYS and forces}
\end{figure}

As shown in Fig. \ref{fig: prototype}(b), MicroASV intakes water from its four inlets at the bottom and propels water out of the four outlet channels at its corners. 
Therefore, the four outlet channels provide thrust forces, while the four inlets are subject to downward forces. However, the forces from the inlets are small and balanced by the buoyancy of MicroASV, and thus the yawing and pitching effects from the downward forces are negligible. Therefore, the system can be described using 3 unconstrained generalized coordinates, with $q_1 = X_G$, $q_2 = Y_G$, and $q_3 = \theta$, as shown in Fig. \ref{fig: C-SYS and forces}. $X_G$ and $Y_G$ are the coordinates of MicroASV's center of mass (COM) position in the inertial frame; $\theta$ is the angle between the $x$-axis of the body-fixed frame and the $X$-axis of the inertial frame.

The general matrix form of Lagrange's equations of motion for the system is written as  
\begin{equation}
    \bm{M}\ddot{\bm{q}}+\bm{C}\dot{\bm{q}} = \bm{Q}^\text{thr}+\bm{Q}^\text{hydr}+\bm{Q}^\text{drag}
    \label{eq: general form}
\end{equation}
where $\bm{q} = [X_G\quad Y_G\quad\theta]^T$ contains the three generalized coordinates, $\dot{\bm{q}}$ and $\ddot{\bm{q}}$ denote the first and second-order time deriavtives of $\bm{q}$,  $\bm{M} \in \mathbb{R}^{3\times 3}$ is the mass matrix, $\bm{C} \in \mathbb{R}^{3\times 3}$ captures the Coriolis and centrifugal effects, and $\bm{Q}^\text{thr}$, $\bm{Q}^\text{hydr}$, $\bm{Q}^\text{drag} \in \mathbb{R}^{3\times 1}$ are vectors containing generalized forces from thrusters, added mass effect, and water drag, respectively.

\subsection{Mass and Coriolis/Centrifugal Matrices}
Let $\bm{R}_\theta \in \mathbb{R}^{2 \times 2}$ denote the rotation matrix from the inertial frame to the body-fixed frame. Let $\bm{J}_G^v \in \mathbb{R}^{2\times 3}$ and $\bm{J}^\omega \in \mathbb{R}^{1\times 3}$ denote the Jacobian matrices for MicroASV's COM velocity and angular velocity, respectively.

The mass matrix $\bm{M}$ can then be evaluated using
\begin{equation}
    \bm{M} = \bm{J}_G^{v^T}m\bm{J}_G^v+\bm{J}^{\omega^T}I^G_{zz}\bm{J}^\omega
    \label{eq: mass matrix}
\end{equation}
where $m$ is the mass of MicroASV, and $I^G_{zz}$ is the moment of inertia of MicroASV about the $z$-axis.

By inspection, the $\bm{C}$ matrix which captures the Coriolis and centrifugal effects should be zero since the generalized coordinates are all evaluated from a stationary position. This can be proved by implementing the closed-form formula for $\bm{C}$ given $\bm{M}$:
\begin{equation}
    C_{ij} = \sum_{k=1}^{3} ((\frac{\partial M_{ij}}{\partial q_k}-\frac{1}{2}\frac{\partial M_{jk}}{\partial {q_i}})\dot{q}_k)
    \label{eq: Coriolis/centrifugal matrix}
\end{equation}
where $M_{ij}$ and $C_{ij}$ denote the elements in the $i$-th row and $j$-th column in the $\bm{M}$ and $\bm{C}$ matrices, and $k$ is a summation index.

\subsection{Generalized Forces from Thrusters}
The numbers and directions of the four thruster forces are shown in Fig. \ref{fig: C-SYS and forces}, and the angle magnitudes between the thrust forces and the $x$-axis are all $\beta$. The four thruster outlets are marked as A, B, C, and D. Computing the Jacobian matrices of the four points helps with finding the generalized forces from the thrusters. The position $\bm{r}_A$ of A expressed in the inertial frame is
\begin{align}
    \bm{r}_A = \bm{R}_\theta
               \begin{bmatrix}
                   -L/2 \\
                   L/2
               \end{bmatrix}
               +
               \begin{bmatrix}
                   X_G \\
                   Y_G
               \end{bmatrix}
\end{align}
Then the Jacobian matrix $\bm{J}_A$ for point A is
\begin{align}
    \bm{J}_A 
             & = \begin{bmatrix}
                    1 & 0 & (\sin(\theta)-\cos(\theta))L/2 \\
                    0 & 1 & -(\sin(\theta)+\cos(\theta))L/2
             \end{bmatrix}
\end{align}
The same procedures can be applied to find $\bm{J}_B$, $\bm{J}_C$, and $\bm{J}_D$.

The generalized forces $\bm{Q}_A^\text{thr}$ of thruster A can be obtained by
\begin{align}
    \bm{Q}_A^\text{thr} = \bm{J}_A^T\bm{R}_\theta
                          \begin{bmatrix}
                          -F_1\cos(\beta) \\
                          -F_1\sin(\beta)
                          \end{bmatrix}
\end{align}
Then $\bm{Q}_B^\text{thr}$, $\bm{Q}_C^\text{thr}$, and $\bm{Q}_D^\text{thr}$ can be obtained by the same procedure, and the total generalized forces $\bm{Q}^\text{thr}$ is
\begin{align}
    \bm{Q}^\text{thr} = \bm{Q}_A^\text{thr}+\bm{Q}_B^\text{thr}
                        +\bm{Q}_C^\text{thr}+\bm{Q}_D^\text{thr}
\end{align}

\subsection{Hydrodynamic Added Mass}
The forces (expressed in the body-fixed frame) and $z$-direction moment from the hydrodynamic added mass effect, $\bm{F}^\text{hydr,bf}$ and $\tau^\text{hydr}$, is calculated using the added mass matrix $\bm{M}_A$, MicroASV's acceleration $\bm{a}_G$, and its $z$-direction angular acceleration $\alpha$, written as
\begin{align}
    \begin{bmatrix}
        \bm{F}^\text{hydr,bf} \\
        \tau^\text{hydr}
    \end{bmatrix}
    =
    -\bm{M}_A
    \begin{bmatrix}
        \bm{R}_\theta^T\bm{a}_G \\
        \alpha
    \end{bmatrix}
\end{align}

For the hydrodynamics part, MicroASV is approximated as a hemisphere, and thus the added mass matrix is approximated as the following diagonal matrix:
\begin{align}
    \bm{M}_A = 
    \rho_\text{water}
    \begin{bmatrix}
        \frac{4}{3}\pi R_\text{eff}^3 & 0 & 0\\
        0 & \frac{4}{3}\pi R_\text{eff}^3 & 0\\
        0 & 0 & \frac{1}{10}\pi R_\text{eff}^5
    \end{bmatrix}
    \label{eq: added mass matrix}
\end{align}
where $R_\text{eff}$ is the effective radius of MicroASV. Then the generalized forces $\bm{Q}^\text{hydr}$ from the hydrodynamic added mass is
\begin{align}
    \bm{Q}^\text{hydr} = \bm{J}_G^{v^T}\bm{R_\theta}\bm{F}^\text{hydr,bf}
                         +\bm{J}^{\omega^T}\tau^\text{hydr}
    \label{eq: Q_hydr}
\end{align}

\subsection{Drag Matrix}
The drag is modeled as linear damping since MicroASV moves at low speeds. Similar to the derivations for the hydrodynamic added mass, the water drag forces (expressed in body-fixed frame) and moment on MicroASV, $\bm{F}^\text{drag,bf}$ and $\tau^\text{drag}$ , is calculated using
\begin{equation}
    \begin{bmatrix}
        \bm{F}^\text{drag,bf} \\
        \tau^\text{drag}
    \end{bmatrix}
    =
    -\bm{D}
    \begin{bmatrix}
        \bm{R}_\theta^T\bm{v}_G \\
        \omega
    \end{bmatrix}
\end{equation}
where $\bm{D}$ is the drag matrix, $\bm{v}_G$ is the velocity of MicroASV's COM, and $\omega$ is the $z$-directional angular velocity of MicroASV. Approximating MicroASV as a hemisphere, the drag matrix is approximated as
\begin{equation}
    \bm{D} = 
    \mu_\text{water}
    \begin{bmatrix}
        4\pi R_\text{eff} & 0 & 0\\
        0 & 4\pi R_\text{eff} & 0\\
        0 & 0 & 0.04\pi R_\text{eff}^2
    \end{bmatrix}
\label{eq: drag matrix}
\end{equation}
where $\mu_\text{water}$ is the viscosity of water. Then the generalized forces $\bm{Q}^\text{drag}$ from the drag is
\begin{align}
    \bm{Q}^\text{drag} = \bm{J}_G^{v^T}\bm{R_\theta}\bm{F}^\text{drag,bf}
                         +\bm{J}^{\omega^T}\tau^\text{drag}
\end{align}

\subsection{Final Form of Dynamic Model}
The generalized forces from hydrodynamic added mass (Eq.(\ref{eq: Q_hydr})) include second-order terms of the generalized coordinates that appear on the right side of the equations of motion. To maintain the state space form of the equations, the second-order time derivative terms in $\bm{Q}^\text{hydr}$ are incorporated into the mass matrix. After this manipulation, the effective mass matrix, $\bm{M}_\text{eff}$, becomes
\begin{equation}
    \bm{M}_\text{eff} = 
    \bm{M}+\bm{M}_A
    \label{eq: effective mass matrix}
\end{equation}
and the final form of the equations of motion, after substituting in $M_\text{eff}$, is
\begin{equation}
    \ddot{\bm{q}} = \bm{M}_\text{eff}^{-1}(\bm{Q}^\text{thr}+\bm{Q}^\text{drag}-\bm{C}\dot{\bm{q}})
    \label{eq: eom final form}
\end{equation}
from which we can obtain the state space form of the governing equations:
\begin{align}
    \dot{\bm{x}} = \bm{f}(\bm{x},\bm{u})
    \label{eq: state space form}
\end{align}
where \(\bm{x} = [\bm{q}^T, \dot{\bm{q}}^T]^T \in \mathbb{R}^{n_{\bm{q}}}\) represents the state vector, \(\quad \bm{u} = [F_1, F_2, F_3, F_4]^T \in \mathbb{R}^{n_{\bm{u}}}\) represents the control vector.
 The system model \(\bm{f}: \mathbb{R}^{n_{\bm{q}} + n_{\bm{u}}} \to \mathbb{R}^{n_{\bm{q}}}\) describes how the state \(\bm{q}\) evolves based on the control input \(\bm{u}\).

\section{Efficient Data-Driven Optimal Control}
\label{sec: control}

\begin{figure*}[!htb]
    \centering
    \includegraphics[width=0.9\textwidth]{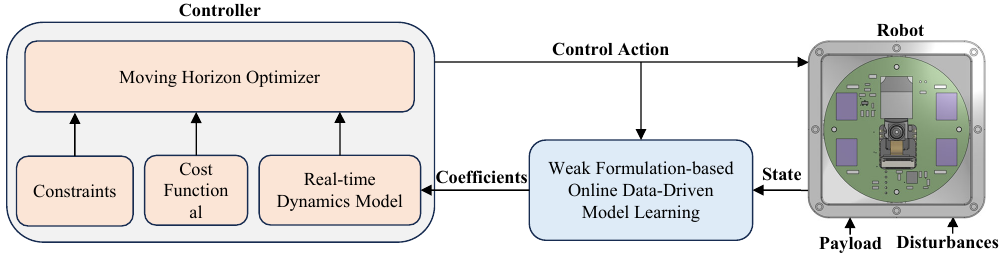}
    \caption{Data-driven optimal control framework for MicroASVs.}
    \label{fig: pipeline}
\end{figure*}

Building on the physics-driven dynamics model, this section introduces a data-driven optimal control framework integrating efficient online model learning with real-time control optimizations, as shown in Fig. \ref{fig: pipeline}.
The weak formulation-based online dynamics learning algorithm robustly refines the dynamics model, which then enables an optimal control strategy tailored for precise and robust trajectory tracking while maintaining computational efficiency suitable for MicroASV.
\subsection{Problem Setup}
The objective of trajectory tracking requires the incorporation of a control law that steers MicroASV to converge to a time-parametrized reference trajectory, minimizing the difference between the desired states and the real states. For this purpose, the reference state as a function of $t$ is defined as:
\begin{equation}  
\bm{x}_d(t) = 
\begin{bmatrix}  
X_d(t)\!\! &\!\!  Y_d(t)\!\!  & \!\! \theta_d(t)\!\!  & \!\! \dot{X}_d(t) \!\! &\!\!  \dot{Y}_d(t)\!\!  &\!\!  \dot{\theta}_d(t) 
\end{bmatrix}^T
\label{eq:reference_state}  
\end{equation} 
where the subscript $d$ denotes a reference. Thus, the controller should minimize the instantaneous tracking error:
\begin{equation}
    e_\text{traj}(t) = \|\bm{x}_d(t)-\bm{x}(t)\|_2
\end{equation}
\subsection{Weak Formulation-Based Online Dynamics Learning}
\label{sec: identification}

A weak formulation-based approach uses integration by parts to bypass direct approximations of higher-order derivatives in the sensor data \cite{messenger2024,bortz2023}. Specifially,
our method for online data-driven modeling essentially sees the dynamics function we aim to identify (i.e., right side of Eq. (\ref{eq: state space form})) as a linear combination of basis functions, where basis functions here refer to a set of functions that span a subspace in a function space, analogous to how basis vectors span a subspace in a vector space. The basis functions consist of state variables in $\bm{x}$, control inputs in $\bm{u}$, and their cross-terms. We first let MicroASV collect trajectory and control input data from $t_a$ to $t_b$, and then we formulate and solve a least-squares problem to find the coefficients of the basis functions. 

For MicroASV, though the numeric values in the $\bm{M_\text{eff}}$ and $\bm{D}$ matrices are not known exactly, the basis functions are known from the physics-driven modeling in Section \ref{sec: physics-driven model}, and the components of $\bm{M_\text{eff}}$ and $\bm{D}$ matrices show as coefficients of the basis functions.

We focus on learning the equations of motion of MicroASV (Eq. (\ref{eq: eom final form})); i.e., we try to identify
\begin{align}
    \ddot{\bm{q}} = \bm{g}(\bm{x},\bm{u})
    \label{eq: eom state space}
\end{align}
where $\bm{g}: \mathbb{R}^{6+4} \rightarrow \mathbb{R}^6$ is the compactly-written form of Eq. (\ref{eq: eom final form}) to facilitate the following derivations. We now focus on learning its first row (the other two rows can be learned using the same approach), denoted as 
\begin{align}
    \ddot{X}_G = g_1 (\bm{x},\bm{u})
    \label{eq: g1}
\end{align}
 According to the physics-driven modeling in Section \ref{sec: physics-driven model}, the basis functions for $g_1$ are $\dot{X}_G$, $F_i \sin(\theta)$, and $F_i \cos(\theta)$, where $i = 1,2,3,4$; i.e.,
\begin{align}
    g_1 (\bm{x},\bm{u}) & = w_1 \dot{X}_G + w_2 F_1 \sin(\theta) \ldots                              +w_9F_4\cos(\theta)
\end{align}
where $w_i$s are the coefficients we aim to identify. 

To implement our weak formulation-based identification approach, we define a set of test functions, $\phi_m(t)$, with compact supports:
\begin{align}
    \phi_m(t)
    =
    \begin{cases}
        C(t-t_a)^p(t_b-t)^q \quad t \in [t_a,t_b] \\
        0   \quad \text{otherwise}
    \end{cases}
\end{align}
where the selection methods for $C$, $p$, and $q$ can be found in \cite{messenger2021}. For each test function $\phi_m$, we multiply it by both sides of Eq. (\ref{eq: g1}), and integrate over the time duration of the collected trajectory data:
\begin{align}
    \int_{t_a}^{t_b} g_1(\bm{x},\bm{u})\phi_m dt
        = \int_{t_a}^{t_b} \ddot{X}_G\phi_m dt
    \label{eq: integration of eom}
\end{align}
Integration by parts transfers the time derivative on the right side of Eq. (\ref{eq: integration of eom}) to the test function:
\begin{align}
    \int_{t_a}^{t_b} g_1(\bm{x},\bm{u})\phi_m dt
        = \dot{X}_G\phi_m \bigg|_{t_a}^{t_b}                         
          -\int_{t_a}^{t_b} \dot{X}_G \dot{\phi}_m dt
    \label{eq: eom integration by parts}
\end{align}
Since the test functions have compact supports, $\phi_m$s are zero at $t_0$ and $t_1$, and thus Eq. (\ref{eq: eom integration by parts}) reduces to the weak derivative form:
\begin{align}
    \int_{t_a}^{t_b} g_1(\bm{x},\bm{u})\phi_m dt
        = -\int_{t_a}^{t_b} \dot{X}_G \dot{\phi}_m dt
    \label{eq: weak derivative}
\end{align}

To numerically evaluate this integral, we first build a library $\bm{\Theta}_1 \in \mathbb{R}^{N\times 9}$ containing the evaluations of the nine basis functions of $g_1$ at $N$ time steps:
\begin{align}
    \bm{\Theta}_1
    =
    \begin{bmatrix}
        \dot{X}_G(t_a) & F_1 \sin\theta(t_a) & \cdots
            & F_4 \cos\theta(t_a) \\
        \dot{X}_G(t_{a+1}) & F_1 \sin\theta(t_{a+1}) & \cdots
            & F_4 \cos\theta(t_{a+1}) \\
        \vdots         & \vdots    & \ddots & \vdots \\
        \dot{X}_G(t_b) & F_1 \sin\theta(t_b) & \cdots
            & F_4 \cos\theta(t_b)
    \end{bmatrix}
\end{align}
In addition, we let $\bm{\Phi_m} \in \mathbb{R}^{1 \times N}$ represent the time series of the test function $\phi_m$, then the integral in Eq. (\ref{eq: weak derivative}) can be numerically evaluated using the trapezoidal rule as
\begin{align}
    \Delta t \bm{\Phi_m} \bm{\Theta}_1 \bm{w}_1 = -\Delta t \dot{\bm{\Phi}}_m
        \dot{\bm{X}}_G
    \label{eq: lstsqr single test function}
\end{align}
where $\bm{w}_1 \in \mathbb{R}^{9\times 1}$ is the vector containing the coefficients of the nine basis functions in $g_1$, and $\Delta t$ represents the step sizes of timesteps (note that we do not let $\Delta t$ cancel out here because the timesteps between datapoints are not necessarily constant, and the actual trapezoid integral involves taking the average of the left and right Riemann sums). Let $\bm{\Phi} \in \mathbb{R}^{M \times N}$ represent the time series for all $M$ test functions (i.e., we stack the time serieses of the test functions into a single matrix $\bm{\Phi}$). Since Eq. (\ref{eq: lstsqr single test function}) is true for all $\bm{\Phi_m}$, then
\begin{align}
    \Delta t \bm{\Phi} \bm{\Theta}_1 \bm{w}_1 = -\Delta t \dot{\bm{\Phi}} \dot{\bm{X}}_G
\end{align}
We can then formulate and solve a least-squares problem for $\bm{w}_1$ aiming to minimize $\|\Delta t \bm{\Phi} \bm{\Theta}_1 \bm{w}_1+\Delta t \dot{\bm{\Phi}} \dot{\bm{X}}_G\|_2^2$. The same approach can be implemented to discover the other two equations of motion by finding $\bm{w}_2$ and $\bm{w}_3$.

\subsection{Data-Driven Optimal Control Law}


Given the identified system dynamics in the form of Eq. (\ref{eq: state space form}), we can formulate an optimal control scheme aiming to minimize a cost functional $J$ subject to the constraints of system dynamics:
\begin{align}
    \bm{x}^*,\bm{u}^* 
    = 
    \underset{\bm{x},\bm{u}}{\text{argmin}}~ J[\bm{x},\bm{u}]
    \quad \text{s.t.}~
    \dot{\bm{x}} = \bm{f}(\bm{x},\bm{u})
\end{align}
where the dynamics constraint, $\dot{\bm{x}} = \bm{f}(\bm{x},\bm{u})$, is learned online from near real-time historal data using a weak formulation-based online dynamics learning algorithm (discussed in the last subsection).
We define the cost functional to be similar to one in an LQR, but we include in the cost the error from dynamic trajectory tracking (a standard LQR regulates the states to zero):
\begin{align}
    J & = \int_{t_0}^{t_f} \biggl(
          \underbrace{\frac{1}{2}(\bm{x}-\bm{x}_d)^T\bm{Q}(\bm{x}-     
          \bm{x_d})+\frac{1}{2}\bm{u}^T
          \bm{R}\bm{u}}_\mathcal{L} \biggr)dt \nonumber \\
          &\qquad 
          +\frac{1}{2}(\bm{x}(t_f)
          -\bm{x}_d(t_f))^T
          \bm{Q}_f(\bm{x}(t_f)-\bm{x}_d(t_f))
\end{align}
where $\bm{x}$ is the state space of the system, $\bm{x}_d$ is the desired trajectory, $t_0$ and $t_f$ are the initial and terminal time instants, and $\bm{Q}$, $\bm{R}$, and $\bm{Q}_f$ are the symmetric positive definite weight matrices. The quadratic forms and the positive definite weight matrices make this optimization problem convex. Thus, a stationary point for $J$ implies a minimum, and the calculus of variations is applicable.

We incorporate constraints using the Lagrange multiplier method, obtaining the augmented cost functional:
\begin{align}
    J_\text{aug} & = \int_{t_0}^{t_f} \biggl(
          \mathcal{L} \nonumber
          +\bm{\lambda}^T(\bm{f}(\bm{x},\bm{u})-\dot{\bm{x}})\biggr)dt
          \nonumber \\
          & \qquad 
          +\frac{1}{2}(\bm{x}(t_f)
          -\bm{x}_d(t_f))^T         
          \bm{Q}_f(\bm{x}(t_f)-\bm{x}_d(t_f))
\end{align}
where $\bm{\lambda} \in \mathbb{R}^{6\times 1}$ represents the Lagrange multipliers enforcing the constraints of system dynamics. 

Taking the variation of $J_\text{aug}$ denoted as $\delta J_\text{aug}$, applying the chain rule, and letting $\delta J_\text{aug}$ be zero yields
\begin{align}
    \delta J_\text{aug} & = \int_{t_0}^{t_f}\biggl(
                 \frac{\partial\mathcal{L}}
                 {\partial\bm{x}}\delta\bm{x}+\frac{\partial\mathcal{L}}
                 {\partial\bm{u}}\delta\bm{u}+\bm{\lambda}^T
                 \frac{\partial \bm{f}}{\partial\bm{x}}\delta\bm{x}+\bm{\lambda}^T
                 \frac{\partial \bm{f}}{\partial\bm{u}}\delta\bm{u} \nonumber \\
                 &~~~ 
                 -\bm{\lambda}^T\delta\dot{\bm{x}}\biggl)dt
                                  +\frac{1}{2}\frac{d}{d\bm{x}(t_f)}\biggl((\bm{x}(t_f)
                 -\bm{x}_d(t_f))^T\bm{Q}_f \nonumber \\
                 &\qquad \qquad \qquad 
                 (\bm{x}(t_f)-\bm{x}_d(t_f))\biggl)
                 \delta\bm{x}(t_f) = 0
    \label{eq: delta J calc.}
\end{align}
The $\delta\dot{\bm{x}}$ term can be converted into $\delta{\bm{x}}$ with integration by parts:
\begin{align}
    \int_{t_0}^{t_f}\underbrace{\bm{\lambda}^T}_u
    \underbrace{\delta\dot{\bm{x}}dt}_{dv}
        & = \bm{\lambda}^T\delta\bm{x}\bigg|_{t_0}^{t_f}
            -\int_{t_0}^{t_f}\dot{\bm{\lambda}}^T\delta\bm{x}dt
            \nonumber \\
        & = \bm{\lambda}(t_f)^T\delta\bm{x}(t_f)
            -\bm{\lambda}(t_0)^T\underbrace{\delta\bm{x}(t_0)}_0 \nonumber \\
            & \qquad \qquad \qquad \quad
            -\int_{t_0}^{t_f}\dot{\bm{\lambda}}^T\delta\bm{x}dt
\end{align}
Therefore, after evaluating the derivatives of the quadratic forms, Eq. (\ref{eq: delta J calc.}) becomes
\begin{align}
    & \int_{t_0}^{t_f}(
    \dot{\bm{\lambda}}^T+\bm{\lambda}^T
    \frac{\partial \bm{f}}{\partial\bm{x}}\delta\bm{x}
    +(\bm{x}-\bm{x}_d)^T\bm{Q}
    )\delta\bm{x}dt \nonumber \\
    &
    +\int_{t_0}^{t_f}(
    \bm{u}^T\bm{R}+\bm{\lambda}^T\frac{\partial \bm{f}}{\partial\bm{u}}
    )\delta\bm{u}dt
    \nonumber \\
    & +\biggl((\bm{x}(t_f)-\bm{x}_d(t_f))^T\bm{Q}_f
    -\bm{\lambda}(t_f)^T\delta\bm{x}(t_f)\biggl)\delta\bm{x}(t_f) = 0
\end{align}
Then the fundamental lemma of the calculus of variations yields
\begin{align}
    & \dot{\bm{\lambda}}^T+\bm{\lambda}^T
    \frac{\partial \bm{f}}{\partial\bm{x}}
    +(\bm{x}-\bm{x}_d)^T\bm{Q}= 0 \label{eq: co-state ode}\\
    & \bm{u}^T\bm{R}+\bm{\lambda}^T\frac{\partial \bm{f}}{\partial\bm{u}} = 0 \label{eq: lambda vs. u}\\
    & (\bm{x}(t_f)-\bm{x}_d(t_f))^T\bm{Q}_f-\bm{\lambda}^T(t_f) = 0 \label{eq: terminal condition}
\end{align}

The controller inputs $\bm{u}$ can be written in terms of $\bm{\lambda}$ using Eq. (\ref{eq: lambda vs. u}), and therefore Eq. (\ref{eq: co-state ode}) and the system dynamics $\dot{\bm{x}} = f(\bm{x},\bm{u})$ forms a system of coupled ODEs for $\bm{x}$ and $\bm{\lambda}$, subject to the boundary conditions $\bm{x}(t_0) = \bm{x}_0$ and $\bm{\lambda}(t_f)-\bm{Q}_f^T(\bm{x}(t_f)-\bm{x}_d(t_f)) = 0$ according to Eq. (\ref{eq: terminal condition}). This is a TPBVP, and we solve it numerically in MATLAB using \texttt{bvp4c()}.

With the numerical solution to the TPBVP as $\bm{x}(t)$ and $\bm{\lambda}(t)$, an optimal control path $\bm{u}^*(t)$ can be obtained using Eq. (\ref{eq: lambda vs. u}). Since the real controller operates on discrete time steps, $\bm{u}^*(t)$ can be linearly interpolated to obtain $\bm{u}^*_k$ at each time step $k$. The optimization is carried out every \SI{1}{\second} (i.e., $t_f-t_0 = \SI{1}{\second}$ for each controller iteration of MicroASV).

\subsection{Algorithm Implementation}
Our MATLAB runs on a Dell G15 laptop with a 13th Gen Intel(R) Core(TM) i7-13650HX 2.60 GHz CPU and 16 GB RAM. We let our controller operate at \SI{100}{\hertz}, and thus 100 $\bm{u}_k$s are calculated each second. Our TPBVP solver for optimal control has a grid density of 20 per second; i.e., among the 100 $\bm{u}_k$s per second, only 20 of them are obtained from solving the TPBVP, and the rest are obtained via linear interpolations. 
Our simulation tests indicate that each 1-second optimization loop completes in approximately 0.05 seconds, including linear interpolations. Assuming the MicroASVs utilize a Raspberry Pi Compute Module 5 and that a typical Intel Core i7 operates 2 to 5 times faster, the optimization time per calculation on these MicroASVs could range between 0.1 and 0.25 seconds. This is shorter than the tested 1-second optimization loop, suggesting that our optimal controller could be feasible for these MicroASVs. Moreover, MATLAB’s \texttt{bvp4c()}  is designed for general applicability rather than computational efficiency in specialized applications. By developing a dedicated TPBVP solver and incorporating warm-start techniques, the computational cost can be significantly reduced, making it likely that certain STM32 microprocessors (MPUs) could implement this optimal controller on MicroASVs.

\section{Results}
\label{sec: results}

To evaluate the performance of our data-driven optimal control framework, we built a MATLAB-based dynamics simulator for MicroASV using the model in Section \ref{sec: physics-driven model}. In the simulations, we let MicroASV perform trajectory tracking tasks following the parametric curves defined in Section \ref{sec: control} with an unknown payload and/or external disturbances.



\subsection{Target Trajectory for Parametric Curve Tracking}
We let MicroASV track specified parametric curves (a sine curve and a spiral) whose parametric equations are $X_d(t)$ and $Y_d(t)$. The reference velocities, denoted as $\dot{X}_d$ and $\dot{Y}_d$, can then be obtained by taking time derivatives of the parametric equations. Then, the reference heading $\theta_d$ is
\begin{align}
    \theta_d = \text{atan}2(\dot{Y}_d,\dot{X}_d)
\end{align}
The reference angular velocity $\dot{\theta}_d$ can then be obtained by taking the time derivative of $\theta_d$. Therefore, the desired trajectory $\bm{x}_d$ can be obtained as Eq. (\ref{eq:reference_state}).

Our sine curve with amplitude $Y_0$ and frequency $\omega$ has parametric equations defined as
\begin{align}
    & X_d = v_0 t \\
    & Y_d = Y_0 \sin(\omega v_0 t)
\end{align}
where $v_0$ is the reference speed in $X$-direction. We define the parametric equations for our spiral as
\begin{align}
    & X_d = v_0 t-1+\cos(\omega t) \\
    & Y_d = v_0 t+\sin(\omega t)
\end{align}
where $v_0$ is the approximate magnitude of MicroASV's average velocity, and $\omega$ is the approximate spin rate of MicroASV as it moves along the spiral.

\subsection{Data-Driven Dynamics Learning}
In our first simulation experiment, we assessed the performance of our data-driven dynamics learning algorithm. We let MicroASV track a sine curve for 30 s (which is the refresh period of the data-driven model) with a payload of 0.2 kg ($= 0.8m_0$, where $m_0 = 0.25$ kg is the unloaded mass of MicroASV). We set the effective radius of MicroASV as 0.08 m, the moment of inertia about the $z$-axis as \SI{0.0045}{\kilo\gram\meter\squared}, and the distance between adjacent thrusters as 0.025 m. 

The dynamics model learning with 30 s of trajectory and control input data took 0.95 s, and the output coefficients are
\begin{align}
    \scriptsize
    \bm{w}_1
    =
    \begin{bmatrix}
        -0.4139 \\
        0.5833 \\
        -0.5817 \\
        0.5837 \\
        0.5816 \\
       -0.5833 \\
        0.5817 \\
       -0.5836 \\
       -0.5816 
    \end{bmatrix},~
    \bm{w}_2
    =
    \begin{bmatrix}
        -0.3953\\
       -0.5559\\
       -0.5649\\
        0.5587\\
       -0.5626\\
        0.5561\\
        0.5649\\
       -0.5589\\
        0.5626
    \end{bmatrix},~
    \bm{w}_3
    =
    \begin{bmatrix}
        -0.0332\\
        0.0697\\
       -0.1440\\
        0.1438\\
       -0.0694
    \end{bmatrix}
\end{align}
For reference, the ground-truth coefficients, derived by substituting the aforementioned parameters into Eq. (\ref{eq: eom final form}), are
\begin{align}
    \scriptsize
    \bm{w}_1^\text{true}
    =
    \begin{bmatrix}
        -0.4\\
        0.5638\\
        -0.5638\\
        0.5638\\
        0.5638\\
        -0.5638\\
        0.5638\\
        -0.5638\\
        -0.5638
    \end{bmatrix},
    \bm{w}_2^\text{true}
    =
    \begin{bmatrix}
        -0.4\\
       -0.5638\\
       -0.5638\\
        0.5638\\
       -0.5638\\
        0.5638\\
        0.5638\\
       -0.5638\\
        0.5638
    \end{bmatrix},
    \bm{w}_3^\text{true}
    =
    \begin{bmatrix}
        -0.0305\\
        0.1072\\
        -0.1072\\
        0.1072\\
        -0.1072
    \end{bmatrix}
\end{align}
Although there are discrepancies in the coefficients in $\bm{w}_3$, the precision of control will not be essentially affected, because $\bm{w}_3$ represents the moment balance equation, and thus $\bm{w}_3$ and $\bm{w}_3^\text{true}$ will result in very close total moments due to the symmetry in the configuration of MicroASV's thrusters.

\subsection{Curve Tracking Evaluation with a Sudden Payload Shift}
We added a \SI{2}{\kilo\gram} payload at \( t = 30 \) s to assess the ability of our data-driven optimal controller to adapt to unexpected payload changes during both sine and spiral curve tracking.
We define the tracking error $e(t)$ for parametric curve following as
\begin{align}
    e(t) = \sqrt{(X_G(t)-X_d(t))^2+(Y_G(t)-Y_d(t))^2}
\end{align}
where $X_G$ and $Y_G$ are the true COM coordinates, and $X_d$ and $Y_d$ are the desired COM coordinates. 

\begin{figure}[!htb]
    \centering
    \includegraphics[width=0.5\textwidth]{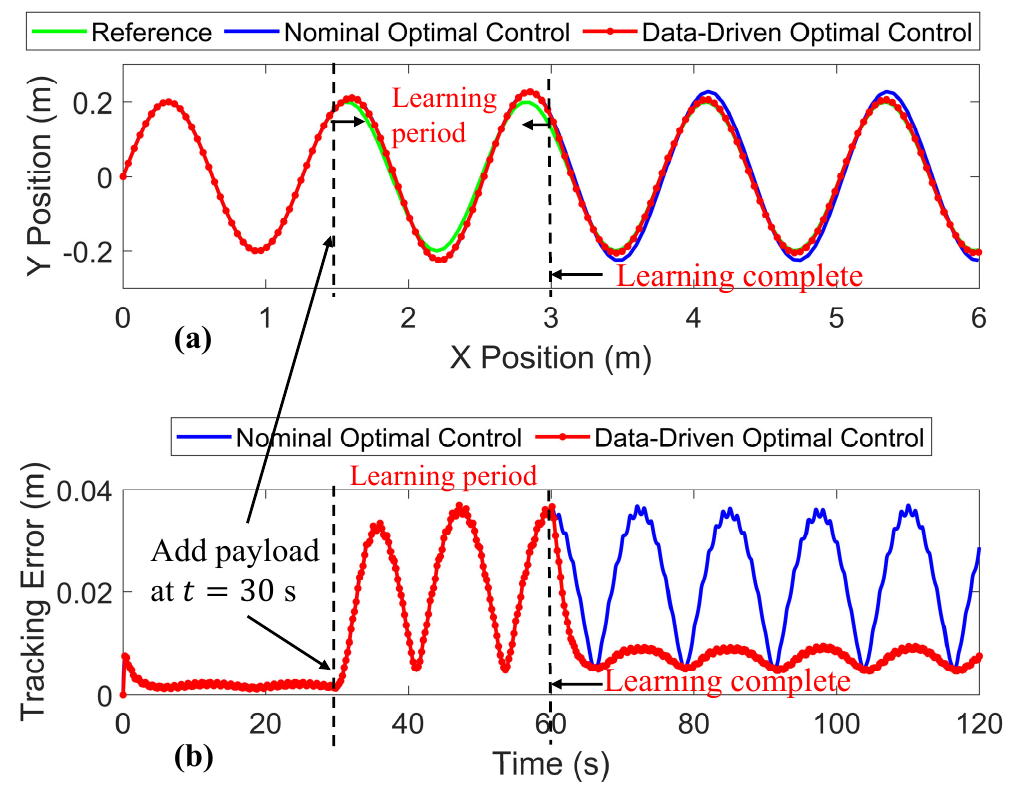}
    \caption{Simulation results of the sine curve tracking test. (a) A plot of the reference path and the paths travelled by MicroASV with nominal and data-driven optimal control. (b) Sine curve tracking error comparison between nominal and data-driven optimal control.}
    \label{fig: sine tracking}
\end{figure}

\begin{figure}[!htb]
    \centering
    \includegraphics[width=0.5\textwidth]{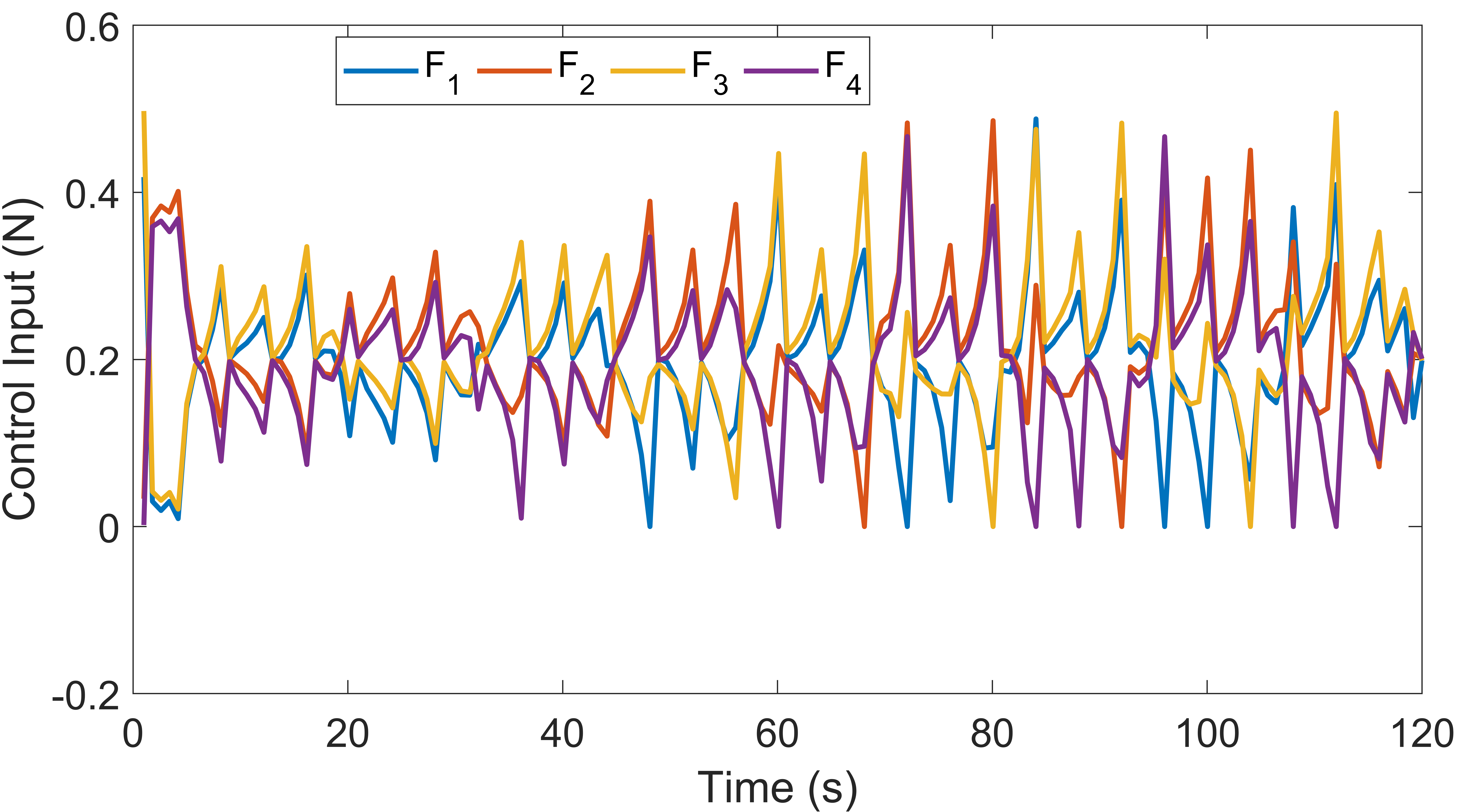}
    \caption{Control inputs of MicroASV thrusters in sine curve tracking test with data-driven optimal control.}
    \label{fig: sine tracking forces}
\end{figure}

 First, we let MicroASV track a sine curve for $120$ seconds. The simulation results for sine curve tracking are shown in Fig. \ref{fig: sine tracking}. 
 The time series for control inputs is shown in Fig. \ref{fig: sine tracking forces}. Note that we use a projection algorithm so that thruster forces are non-negative. This projection algorithm involves solving a constrained underdetermined system with an initial guess of $0.2$ N, which is why the plot for control inputs seems close to being symmetric about $0.2$ N.

It can be seen from Fig. \ref{fig: sine tracking} (a) that switching to the identified model reduces the overshoot of MicroASV at turns of the sine curve. Fig. \ref{fig: sine tracking} (b) shows that when the learning is complete, both the average and maximum tracking error decreases significantly. Numerically, our online identification algorithm reduces the average tracking error by $75.1\%$ and reduces the maximum tracking error by $70.7$\%.

\begin{figure}[!htb]
    \centering
    \includegraphics[width=0.5\textwidth]{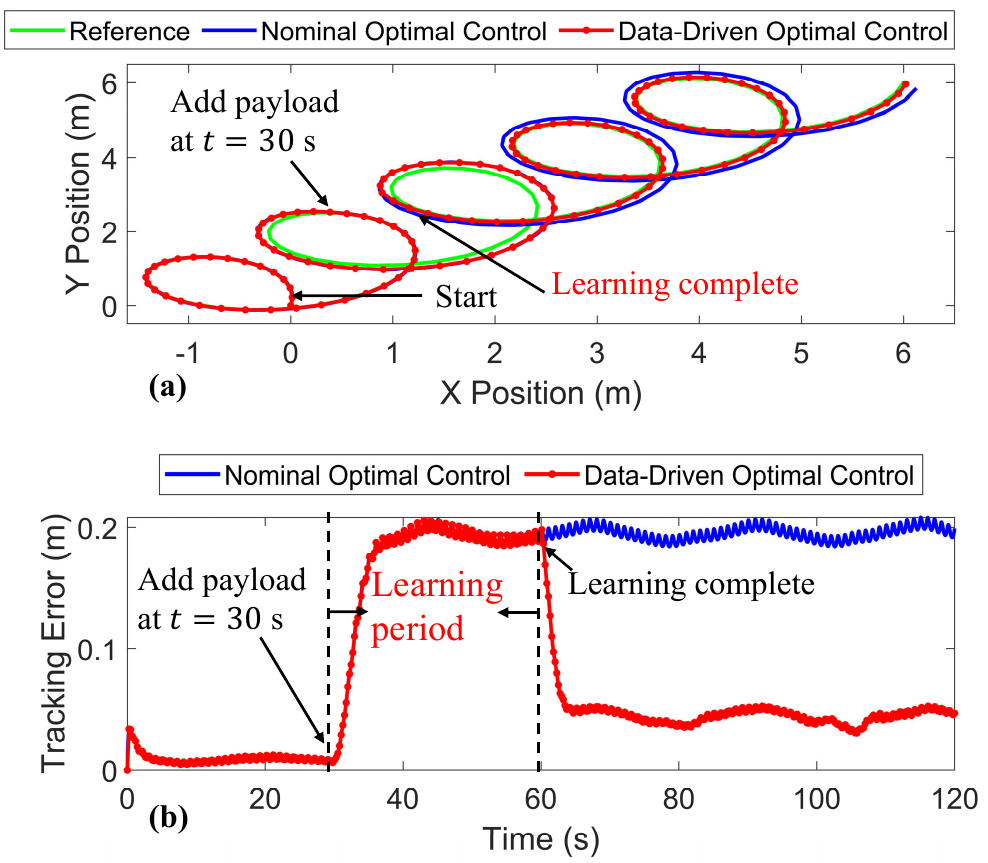}
    \caption{Simulation results of the spiral tracking test. (a) A plot of the reference and actual paths. (b) Spiral tracking error comparison for MicroASV with and without the online identification.}
    \label{fig: spiral tracking}
\end{figure}

Next, we let MicroASV track a spiral for $120$ s. The simulation results are shown in Fig. \ref{fig: spiral tracking}. 
From Fig. \ref{fig: spiral tracking} (a), the data-driven optimal control considerably eliminates the stead-state error of the spiral tracking; Fig. \ref{fig: spiral tracking} (b) confirms that switching to the identified model drastically decreases the tracking error. Numerically, the online identification reduces the average tracking error by $71.4\%$ and reduces the maximum tracking error by $75.7\%$.

\subsection{Tracking Evaluation with a Varying-Weight Payload and Model Inaccuracy}
Real-world applications of robots, such as the MicroASV, often encounter inaccuracies in dynamics and varying payloads. In this subsection, we let the nominal model have inaccuracies in both payloads and a hydrodynamic parameter, the effective radius $R_\text{eff}$, in contrast to the previous subsections where we only assume inaccuracies in payloads for the nominal model. We pick $R_\text{eff}$ because the hemisphere assumption can be subject to error without dedicated CFD simulations. Specifically, we set $R_\text{eff}$ of the nominal model to be 2-cm larger than the ground-truth $R_\text{eff}$ used for the simulation (which is a reasonable assumption given the corners in MicroASV's geometry). First, we set up the MicroASV's controller to use conventional optimal control with the nominal dynamics model; then, we enable the data-driven optimal control, allowing the MicroASV to learn and update the real-time dynamics every 30 seconds. We let MicroASV track a sine curve for $120$ seconds multiple times; each time, MicroASV carried a payload ranging from 0 to \SI{2}{\kilo\gram} ($= 8m_0$) with an interval of 0.2 kg.

\begin{figure}[!htb]
    \centering
    \includegraphics[width=0.46\textwidth]{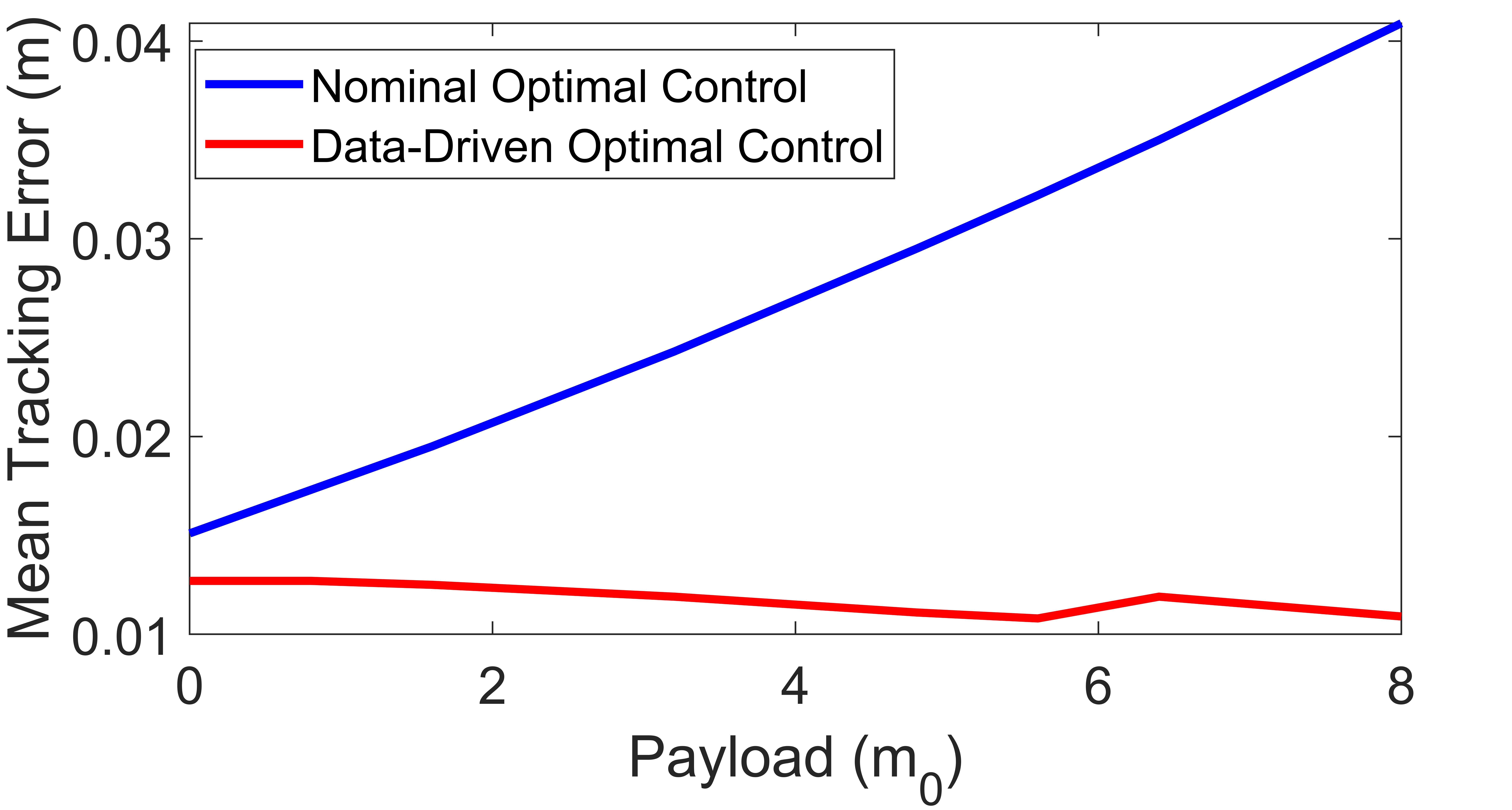}
    \caption{A comparison between the mean tracking error of MicroASV with nominal and data-driven optimal control, as payload increases.}
    \label{fig: increasing payload}
\end{figure}

The mean tracking error comparison between MicroASV with and without online identification, as payload increases, is shown in Fig. \ref{fig: increasing payload}.  At a 2-kg payload with inaccuracies in the nominal model, the mean tracking error with data-driven optimal control decreases by 75\%.

\subsection{Tracking Evaluation With Disturbances}
To demonstrate that the data-driven controller enhances the robustness of the system under disturbances when the controller learns the dynamics in real time, we have the MicroASV track a sine curve for 50 seconds with a 2-kg payload, using two different controllers: one with optimal control based on a nominal model that assumes no payload, and the other with data-driven optimal control that identifies the 2-kg payload and updates the dynamics in real time.  

\begin{figure}
    \centering
    \includegraphics[width=0.46\textwidth]{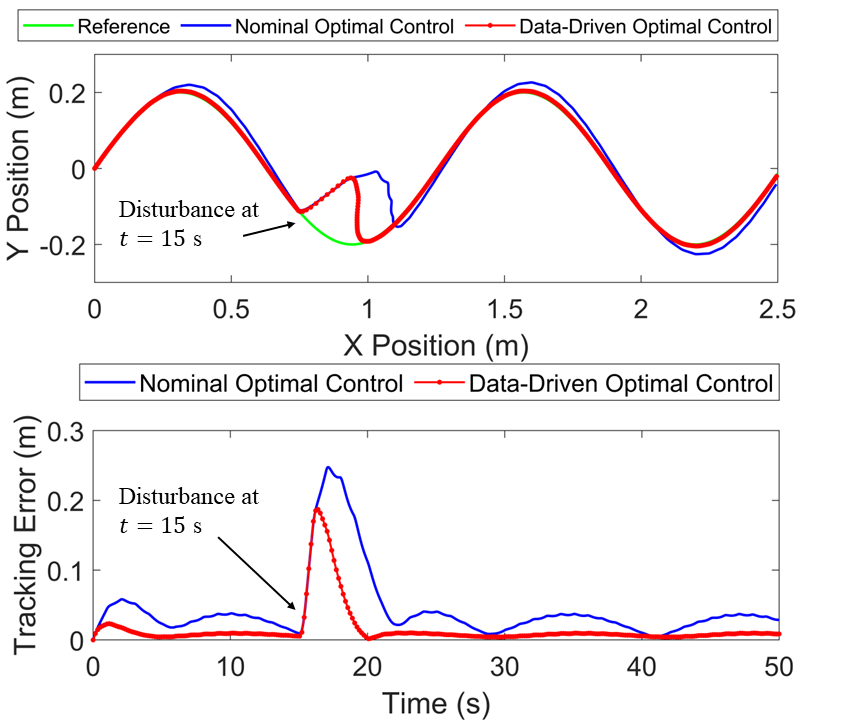}
    \caption{Simulation results of the sine curve tracking test with disturbance occurring at $t = 15$ s. (a) A comparison of the reference path and the path travelled by MicroASV with and without the data-driven model. (b) Tracking error comparison.}
    \label{fig: disturbance}
\end{figure}

In both simulations, a disturbance is introduced at $t =15$ s. The disturbance is modeled as a force-couple system exerted on the COM of MicroASV for a duration of $0.5$ s. The force has magnitudes of $1$ N in both $X$ and $Y$ directions, and the couple has a magnitude of $0.5$ Nm in the $Z$ direction. The comparison between the two trajectory tracking results with disturbance is shown in Fig. \ref{fig: disturbance}.
We can see that the data-driven optimal control that learns the correct dynamics has less overshoot compared to the nominal optimal control when subject to disturbance; in addition, the tracking error of the data-driven controller converges faster to a normal range after the disturbance. Numerically, the data-driven optimal control algorithm decreases the overshoot from disturbance by $23.5\%$, and reduces the convergence time by $35.7\%$.

\section{Conclusion and Future Work}
\label{sec: conslusion}

To conclude, our work develops a physics-driven dynamics model for MicroASV and a data-driven optimal control framework. The framework consists of an online weak formulation-based dynamics learning method that refines the dynamics model iteratively and a TPBVP-based optimal controller that uses the learned dynamics to improve its performance. Simulation results show that our data-driven optimal control reduces trajectory-tracking errors and handles disturbances more effectively, making it a strong choice for precise and efficient MicroASV control.

Our future work includes implementing the data-driven optimal control scheme on the real hardware systems of MicroASV. In addition, we are interested in further improving the accuracy of the online identification outputs and improving the computational speed of our TPBVP-based optimal control algorithm.

\bibliographystyle{IEEEtran}
\bibliography{references}

\end{document}